\def\year{2020}\relax
\documentclass[letterpaper]{article} 
\usepackage{aaai20}  
\usepackage{times}  
\usepackage{helvet} 
\usepackage{courier}  
\usepackage[hyphens]{url}  
\usepackage{graphicx} 
\usepackage{graphicx}  
\usepackage{amsmath,amsfonts,amssymb} 
\usepackage{multirow}  
\usepackage{booktabs} 
\usepackage{subcaption} 
\usepackage{algorithm,algpseudocode} 
\title{An Attention-based Graph Neural Network for\\ Heterogeneous Structural Learning}
\author{Huiting Hong,\textsuperscript{\rm 1}
Hantao Guo,\textsuperscript{\rm 2}\thanks{This work was done when the second author was an intern at AI Labs, Didi Chuxing, Beijing, China.}
Yucheng Lin,\textsuperscript{\rm 1}
Xiaoqing Yang,\textsuperscript{\rm 1}
Zang Li,\textsuperscript{\rm 1}\thanks{Corresponding author.}
Jieping Ye\textsuperscript{\rm 1}\\
\textsuperscript{\rm 1}AI Labs, Didi Chuxing, Beijing, China,
\textsuperscript{\rm 2}Peking University, Beijing, China.\\
honghuiting@didiglobal.com, guohantao@pku.edu.cn, linyucheng@didiglobal.com,\\yangxiaoqing@didiglobal.com, lizang@didiglobal.com, yejieping@didiglobal.com
}
 
\begin{document}

\maketitle

\begin{abstract}
In this paper, we focus on graph representation learning of heterogeneous information network (HIN), in which various types of vertices are connected by various types of relations. Most of the existing methods conducted on HIN revise homogeneous graph embedding models via meta-paths to learn low-dimensional vector space of HIN. In this paper, we propose a novel Heterogeneous Graph Structural Attention Neural Network (HetSANN) to directly encode structural information of HIN without meta-path and achieve more informative representations. With this method, domain experts will not be needed to design meta-path schemes and the heterogeneous information can be processed automatically by our proposed model. Specifically, we implicitly represent heterogeneous information using the following two methods: 1) we model the transformation between heterogeneous vertices through a projection in low-dimensional entity spaces; 2) afterwards, we apply the graph neural network to aggregate multi-relational information of projected neighborhood by means of attention mechanism. We also present three extensions of HetSANN, i.e., voices-sharing product attention for the pairwise relationships in HIN, cycle-consistency loss to retain the transformation between heterogeneous entity spaces, and multi-task learning with full use of information. The experiments conducted on three public datasets demonstrate that our proposed models achieve significant and consistent improvements compared to state-of-the-art solutions.
\end{abstract}

\section{Introduction}\label{sec:intro}
Graph embedding pursues informative numerical representations of graphs, which facilitates various applications on graphs such as classification, link prediction and entity alignment. Most of existing methods perform graph embedding on homogeneous graphs, where all nodes and relationships (a.k.a. linkages or edges) are of the same type. For example, DeepWalk \cite{deepwalk} minimize the distance between the node and its neighboring nodes in the low-dimensional vector space, to preserve the structural information of the homogeneous graph.
However, the real-world data tends to be presented as a heterogeneous graph, which combines different aspects of information together.
\begin{figure}[t]
	\centering
	\subcaptionbox{\label{fig:het_graph}}
	{\includegraphics[width=0.21\textwidth]{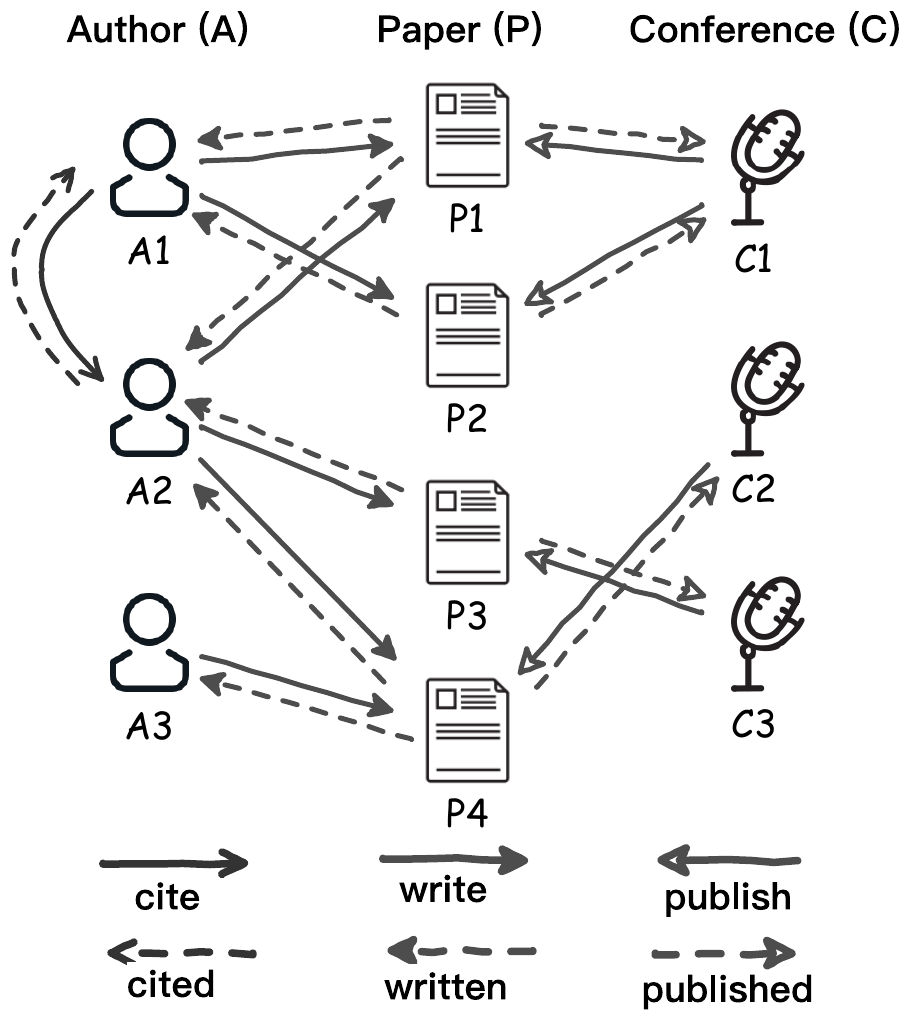}}
	\hskip 5pt
	\subcaptionbox{\label{fig:meta_path}}
	{\includegraphics[width=0.23\textwidth]{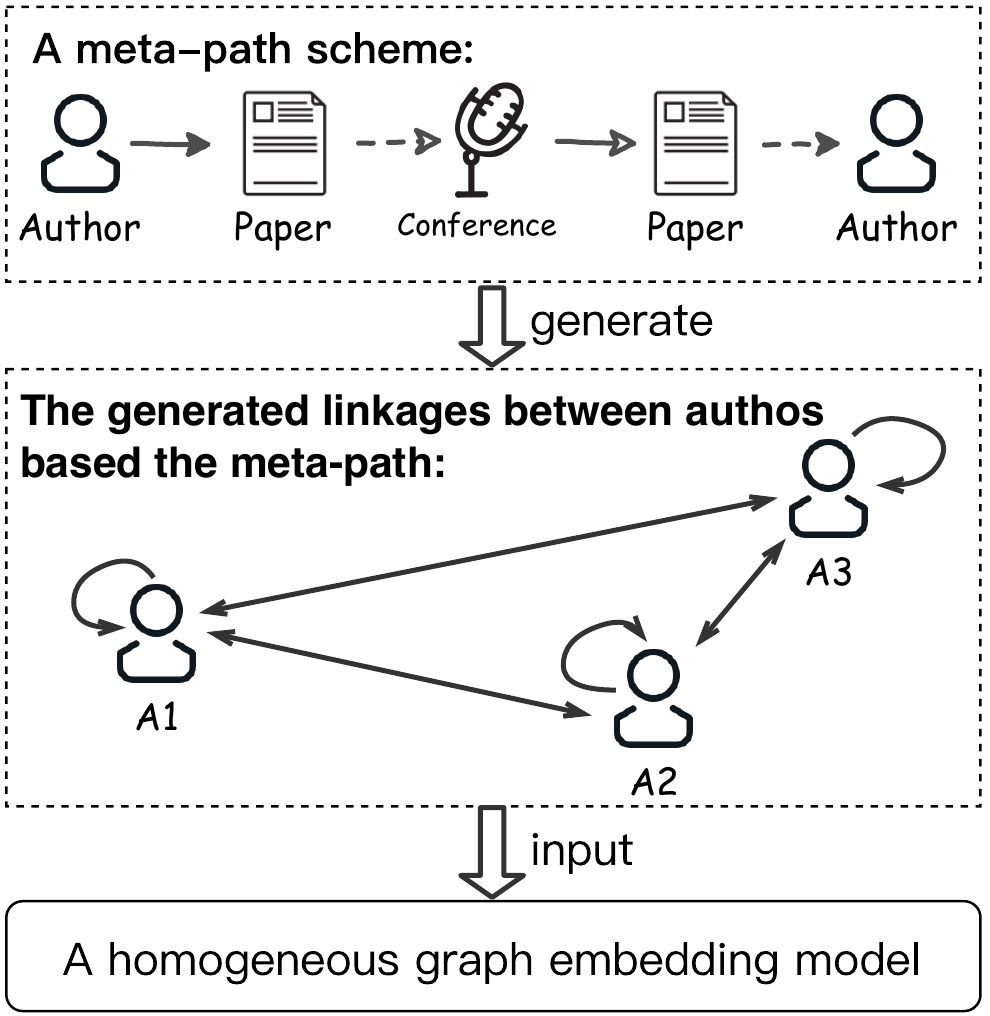}}
	\caption{A toy example of the heterogeneous graph and the meta-path. (a) A heterogeneous academic graph. (b) Previous meta-path-based method.}
	\label{fig:het_graph_example}
\end{figure}
\begin{figure*}[t]
	\centering
	\includegraphics[width=.89\textwidth]{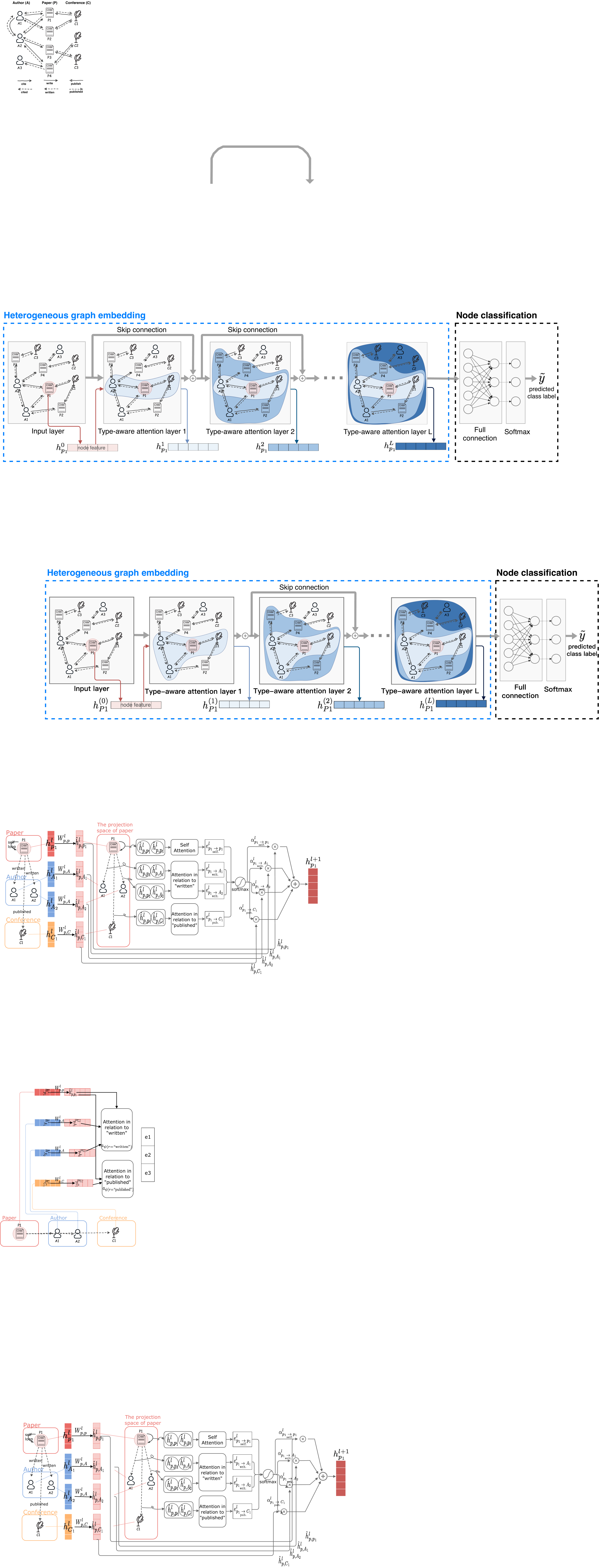}
	\caption{The overall framework of our proposed model HetSANN.}
	\label{fig:hetSANN_overview}
\end{figure*}

\textbf{Heterogeneous Information Network (HIN).} A HIN (a.k.a. heterogeneous graph) comprises more than two types of nodes or edges. Fig. \ref{fig:het_graph} illustrates a toy example of HIN, including three types of nodes (author, paper, and conference) and six types of edges (cite/cited, write/written, and publish/published). Note that here we regard the relationship between vertices in HIN as the directed edge, and we set reverse relationships (e.g., written) for the directed relationships (e.g., write) in HIN. Compared with the homogeneous graph, HIN suffers from two major challenges:
\begin{itemize}
\item \textbf{C1:} how to model the entity space of multiple types of nodes? In a homogeneous graph, all nodes are embedded into the same low-dimensional entity space. In contrast, various types of nodes in HIN are naturally modeled in distinct spaces. However, a vertex may connect to multiple types of nodes, e.g., a paper is written by the author and it will be published by an academic conference. It is imperative to design the way of interaction between vertices in different type-specific entity space.
\item \textbf{C2:} how to preserve the semantic of different relationships between nodes? For a HIN, there exist a variety of relations between both different node pairs and the same node pair. In the case of academic graph, an author can cite another author and meanwhile they can be the co-authors of some paper. The various relationships draw different semantic contents of the vertex. Thus the characterization of neighboring vertices with different relations to a vertex determines the performance of learned low-dimensional representation space.
\end{itemize}
Most of contemporary researches in HIN embedding focus on adapting HIN to homogeneous representation learning algorithms via the meta-path \cite{survey_HIN}. As shown in Fig. \ref{fig:meta_path}, the linkages between authors can be generated based on the designed meta-path scheme \textit{APCPA}, and a representation learning algorithm for homogeneous graphs, e.g., DeepWalk \cite{deepwalk} adopted in metapath2vec \cite{metapath2vec} or GNN \cite{GNN} used in HAN \cite{HAN}, is implemented to the generated graph. See Section \ref{sec:related} for more details on meta-path-based methods.

Despite the success of meta-path-based heterogeneous graph embedding methods, these solutions employ handcrafted meta-path schemes to find homogeneous node neighbors, making them suffer from two predominant problems: 1) the scheme of meta-path relies on experts, and it is hard to exhaustively enumerate and select valuable meta-path schemes by hand; 2) the information passing by the meta-path, such as features of heterogeneous nodes or edges, is lost in the process of generating meta-path based node pairs, and it may even lead to an inferior embedding performance.

In this paper, we cast the meta-path aside and propose a novel method to learn the low-dimensional vector space preserving both structural and semantics information in HIN. Specifically, we take advantage of graph neural network (GNN) to conduct the structural information of HIN and we train the model by the task-guided objective function (node classification loss in this paper). To tackle the challenges of HIN mentioned above, we design a dedicated Type-aware Attention Layer instead of the convolutional layer in the conventional GNN. For each type-aware attention layer, a transformation operation that projects vertices from different entity space to the same low-dimensional target space is defined for the interaction between heterogeneous nodes \textbf{(C1)}, and the attention strategies focusing on different types of edges are applied for the aggregation of neighboring vertices with different semantics \textbf{(C2)}.
Moreover, we develop two kinds of attention scoring functions of proposed type-aware attention layer including \textit{concat product} and \textit{voices-sharing product}\footnote{The voice is the concept of English grammar including active voice and passive voice. Here we refer the active voice to the directed edge (cite, write, etc.) and refer the passive voice to the reversed edge (cited, written, etc.).}. To better model the interaction between heterogeneous nodes, we further introduce a restriction to the transformation operation. Finally, we perform multi-task learning in our proposed model which generally benefits the robustness of representations.
\begin{figure*}[t]
	\centering
	\includegraphics[width=0.913\textwidth]{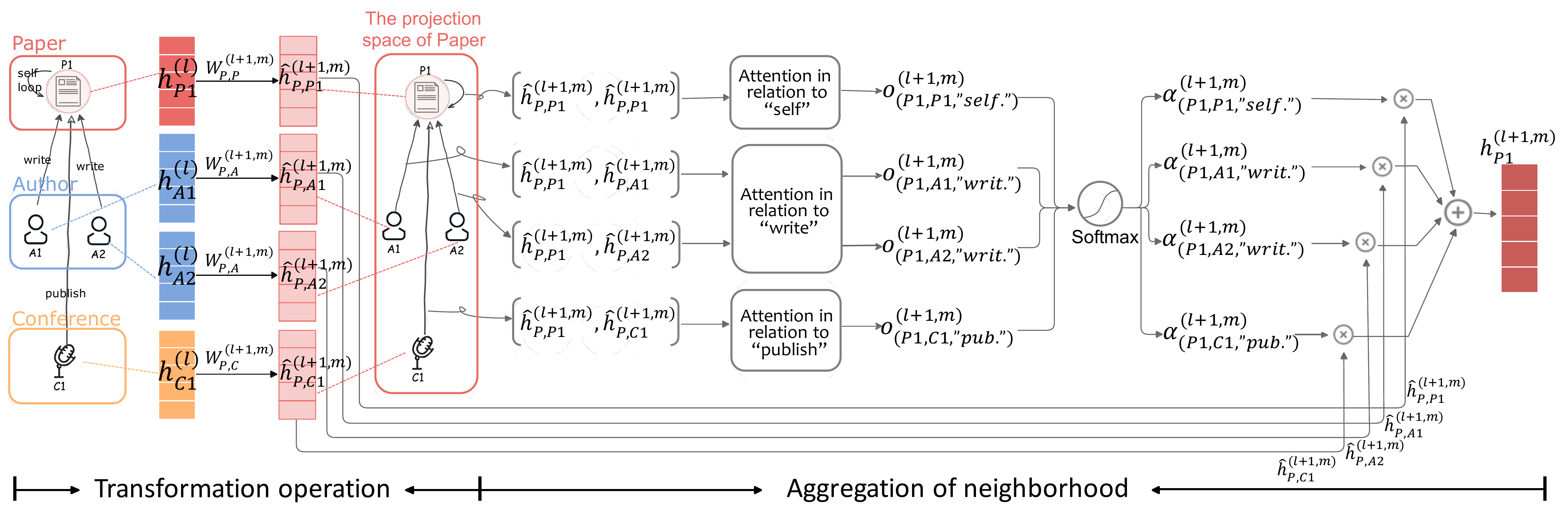}
	\caption{The dataflow of each head of the type-aware attention layer in HetSANN.}
	\label{fig:type_aware_attention}
\end{figure*}
To sum up, the main contributions of this paper are as follows:
\begin{itemize}
	\item We propose \textbf{Het}erogeneous Graph \textbf{S}tructural \textbf{A}ttention \textbf{N}eural \textbf{N}etwork (HetSANN). Unlike previous meta-path-based solutions, HetSANN directly leverages and explores the structures in the heterogeneous graph to achieve more informative representations.
	\item We present three extensions of HetSANN: (E1) Enhance the extent of sharing information with multi-task learning. (E2) Take the pairwise relationship between the directed edge and the reversed edge into account (\textit{voices-sharing product}). (E3) Introduce a constraint to the transformation operation to keep cycle consistent.
	\item We evaluate the proposed HetSANN with the node classification task on three heterogeneous graph datasets. The experimental results demonstrate the superiority of HetSANN compared to various state-of-the-arts. In addition, an ablation study about the three extensions of HetSANN is conducted and the result shows that all three extensions achieve improvement upon the vanilla of HetSANN.
\end{itemize}

\section{Heterogeneous Graph Structural Attention Neural Network (HetSANN)}

A heterogeneous graph $\mathcal{G} = (\mathcal{V}, \mathcal{E})$ consists of a set of vertices $\mathcal{V}$ and a set of edges $\mathcal{E}$. There is a set of node types $\mathcal{A}$, and each vertex $v \in \mathcal{V}$ belongs to one of the node types, denoted by $\phi (v) = p \in \mathcal{A}$, where $\phi$ is the mapping function from $\mathcal{V}$ to $\mathcal{A}$. We represent an edge $e \in \mathcal{E}$ from the vertex $i \in \mathcal{V}$ to $j \in \mathcal{V}$ with a relation type $r$ as a triplet $e = (i, j, r)$, where $r \in \mathcal{R}$ and $\mathcal{R}$ is the set of relation types. For a directed edge $e=(i,j,r)$ in canonical direction, we consider its reversed edge as $\tilde{e}=(j,i,\tilde{r})$ where $\tilde{r}\in \mathcal{R}$ is different from $r$. For a vertex $j$, the set of linkages with its neighboring nodes is defined as $\mathcal{E}_j=\{(i,j,r)\in \mathcal{E}\mid i\in \mathcal{V}, r\in\mathcal{R}\}$. 

In this paper, we aim to learn the low-dimensional representation $\mathbf{h}_i \in \mathbb{R}^{n_{\phi(i)}}$, where $n_{\phi(i)}$ is the dimension of embedding space for node type $\phi(i)$, for each vertex $i$ in the heterogeneous graph and apply it to the downstream node classification task. Note that various relationship types can occur simultaneously when the vertex $i$ links to $j$, which would be a challenge of the heterogeneous graph embedding. To tackle these challenges, we propose a task-guided heterogeneous graph embedding method, namely HetSANN. As shown in Fig. \ref{fig:hetSANN_overview}, the key component of HetSANN framework is the type-aware attention layer presented as follows.

\subsection{Type-aware Attention Layer (TAL)}
The TAL is primarily motivated as an adaptation layer of GNNs, which performs convolution operation on local graph neighborhoods. Before conducting the embedding procedure, we connect each vertex $i$ to itself with the self-loop relation about $\phi (i)$. And we have the cold start state $\mathbf{h}_i^{\left(0\right)} \in \mathbb{R}^{n^{(0)}_{\phi(i)}}$ for per node $i \in \mathcal{V}$. The cold start state can be either the attribute features of nodes, or the dummy features (zero vector/one-hot vector) for the nodes without attributes.

Each TAL employs multi-head attention mechanism \cite{Transformer}, which has been proved that it is helpful to stabilize the learning process of attention mechanism and enrich the model capacity \cite{GAT}. The dataflow of each head of the TAL is illustrated in Fig. \ref{fig:type_aware_attention}. Consider a vertex $j\in \mathcal{V}$ presented as $\mathbf{{h}}_j^{\left(l\right)}$ in the $l$-th layer. An attention head $m$ in the $(l\!+\!1)$-th TAL outputs the corresponding hidden state $\mathbf{{h}}_j^{\left(l+1, m\right)}$ by the following two operations: the transformation operation and the aggregation of the neighborhood in the in-degree distribution of vertex $j$.

\subsubsection{Transformation Operation (C1)}
We first apply a linear transformations $W_{\phi\left(j\right),\phi\left(i\right)}^{\left(l+1, m\right)} \in \mathbb{R}^{n_{\phi\left(j\right)}^{\left(l+1,m\right)}\times n_{\phi\left(i\right)}^{\left(l\right)}}$ to each neighboring vertex $i$ of vertex $j$:
\begin{equation}
\hat{\mathbf{h}}_{\phi\left(j\right), i}^{\left(l+1, m\right)} = W_{\phi\left(j\right),\phi\left(i\right)}^{\left(l+1, m\right)} \mathbf{h}_i^{\left(l\right)}
\label{fun:trans}
\end{equation}
where $\hat{\mathbf{h}}_{\phi\left(j\right), i}^{\left(l+1, m\right)} \in \mathbb{R}^{n_{\phi\left(j\right)}^{\left(l+1,m\right)}}$ is the projection from previous hidden state in the space of type $\phi(i)$ to the hidden space of node type $\phi\left(j\right)$ in $m$-th head of layer $l+1$. That is, we transform the neighboring nodes of vertex $j$ to the same low-dimensional vector space of the node type $\phi(j)$, intended for the neighborhood aggregation.

\subsubsection{Aggregation of Neighborhood (C2)}
To preserve the semantic of different types of relationship between nodes, we utilize $\left|\mathcal{R}\right|$ attention scoring functions to match different relation patterns, i.e., $\mathcal{F}^{(l+1, m)} = \{f_r^{(l+1, m)}\mid r\in \mathcal{R}\}$. For a vertex $j$, an attention coefficient is computed for each link edge $e = \left(i, j, r\right) \in \mathcal{E}_j$ in the form as:
\begin{equation}
o_e^{\left(l+1, m\right)} = \sigma\left(f_r^{\left(l+1,m\right)}\left(\hat{\mathbf{h}}_{\phi\left(j\right), j}^{\left(l+1,m\right)},
\hat{\mathbf{h}}_{\phi\left(j\right), i}^{\left(l+1,m\right)}\right)\right)
\end{equation}
where $\sigma$ is an activation function implemented by $\text{LeakyReLU}(\cdot)$ \cite{leakyrelu}. The attention coefficient $o_e$ indicates the importance of edge $e$ to the target vertex $j$.
In principle, the attention scoring functions can be different forms to capture various link types. For simplicity, we adopt the same form of attention mechanism for all linkage types but different in the parameters. 
A natural form of the attention scoring function is the \textit{concat product}, which is adopted in GAT \cite{GAT}, defined as:
\begin{equation}
f_r^{\left(l+1,m\right)}\left(e\right) =\left[{{{}\hat{\mathbf{h}}}_{\phi\left(j\right), j}^{\left(l+1, m\right)}}^\mathsf{T} \| {{{}\hat{\mathbf{h}}}_{\phi\left(j\right), i}^{\left(l+1,m\right)}}^\mathsf{T} \right]{\mathbf{a}_r^{\left(l+1, m\right)}}
\end{equation}
where $\|$ denotes the concatenation operation, and $\mathbf{a}_r^{(l+1,m)} \in \mathbb{R}^{2n_{\phi\left(j\right)}^{\left(l+1,m\right)}}$ is the trainable attention parameter shared by the same edge type $r$.
Different with the HAN \cite{HAN} which employs the hierarchical attention mechanism based on the meta-path schemes, we utilize the attention mechanism directly to the raw heterogeneous links. Thereby, the softmax is applied over the neighborhood linkages of vertex $j$ for the normalization of the attention coefficient:
\begin{equation}
\alpha_e^{\left(l+1,m\right)} =  {\exp\left(o_e^{\left(l+1,m\right)}\right)}/{\sum_{k \in \mathcal{E}_j}\exp\left(o_k^{\left(l+1,m\right)}\right)}.
\end{equation}

Now we have the hidden states of neighboring nodes in the same low-dimensional space of the target node $j$, and weights of each linkage associated with vertex $j$. Then the neighborhood aggregation for vertex $j$ can be performed as:
\begin{equation}
\mathbf{h}_j^{\left(l+1,m\right)}  = \sigma\Big(\sum_{e = \left(i, j, r\right) \in \mathcal{E}_j} \alpha_e^{\left(l+1,m\right)}\hat{\mathbf{h}}_{\phi\left(j\right), i}^{\left(l+1,m\right)}\Big).
\label{fun:agg_function}
\end{equation}
In our proposed model, the node pair of edge and the relation type of edge are used together to identify edges. When vertex $i$ links to $j$ with multiple types of relationship, the hidden state $\hat{\mathbf{h}}_{\phi\left(j\right), i}^{\left(l+1,m\right)}$ is propagated to vertex $j$ multiple times with the corresponding weight $\alpha_{(i,j,r)}^{(l+1,m)}$.

With $M$ attention heads executing the procedure of Eq. \eqref{fun:agg_function}, we concatenate the low-dimensional vectors of attention heads and output the representation of each node in the type-aware attention layer $l+1$:
\begin{equation}
\mathbf{h}_j^{\left(l+1\right)} = \mathop{\big{\|}}_{m=1}^M \mathbf{h}_j^{\left(l+1, m\right)},
\label{fun:concate_multi_head}
\end{equation}
where $\mathbf{h}_j^{\left(l+1\right)}\in \mathbb{R}^{{\sum}^M_{m=1}n^{(l+1,m)}_{\phi(j)}}$. 
The aggregation of HetSANN is conducted on the raw links instead of the generated links based on meta-paths. That is, a vertex $i$ can be propagated to vertex $j$ within one layer of GNN for the meta-path-based links, while more layers are needed for the raw links. Thus, a deeper model is used in HetSANN to capture the high-order proximity information. To facilitate training, we adopt the residue mechanism, which is first introduced by \cite{residual}, and we revise Eq. \eqref{fun:concate_multi_head} as following:
\begin{equation}
\mathbf{h}_j^{\left(l+1\right)} = \mathbf{h}_j^{\left(l\right)} + \mathop{\big{\|}}_{m=1}^M \mathbf{h}_j^{\left(l+1, m\right)}.
\end{equation}

\subsection{Model Training and Three Extensions}\label{sec:3extension}
The last type-aware attention layer outputs the low-dimensional representations for each vertex in the heterogeneous graph, i.e. $\mathbf{h}_j = \mathbf{h}_j^{\left(L\right)}$. To optimize the representations toward the target task, such as node classification in this paper, we integrate the representations of nodes into a node classifier (implemented with a full connection layer with softmax function) to infer the label of classification. With the guide of labeled data, we minimize the cross-entropy loss:
\begin{equation}
\mathcal{J}_{\text{class}} = -\underset{i\in{\mathcal{V}_p}}{\sum}y_{i}\log \tilde{y}_i,
\end{equation}
where $\mathcal{V}_p$ is the set of labeled vertices belonging to the node type $p$. $y_i$ and $\tilde{y}_i$ are the ground truth and the predicted class label for vertex $i$, respectively.
\begin{figure}[t]
	\centering
	\subcaptionbox{\label{fig:trans_oper}}
	[.48\linewidth]{\includegraphics[width=0.2\textwidth]{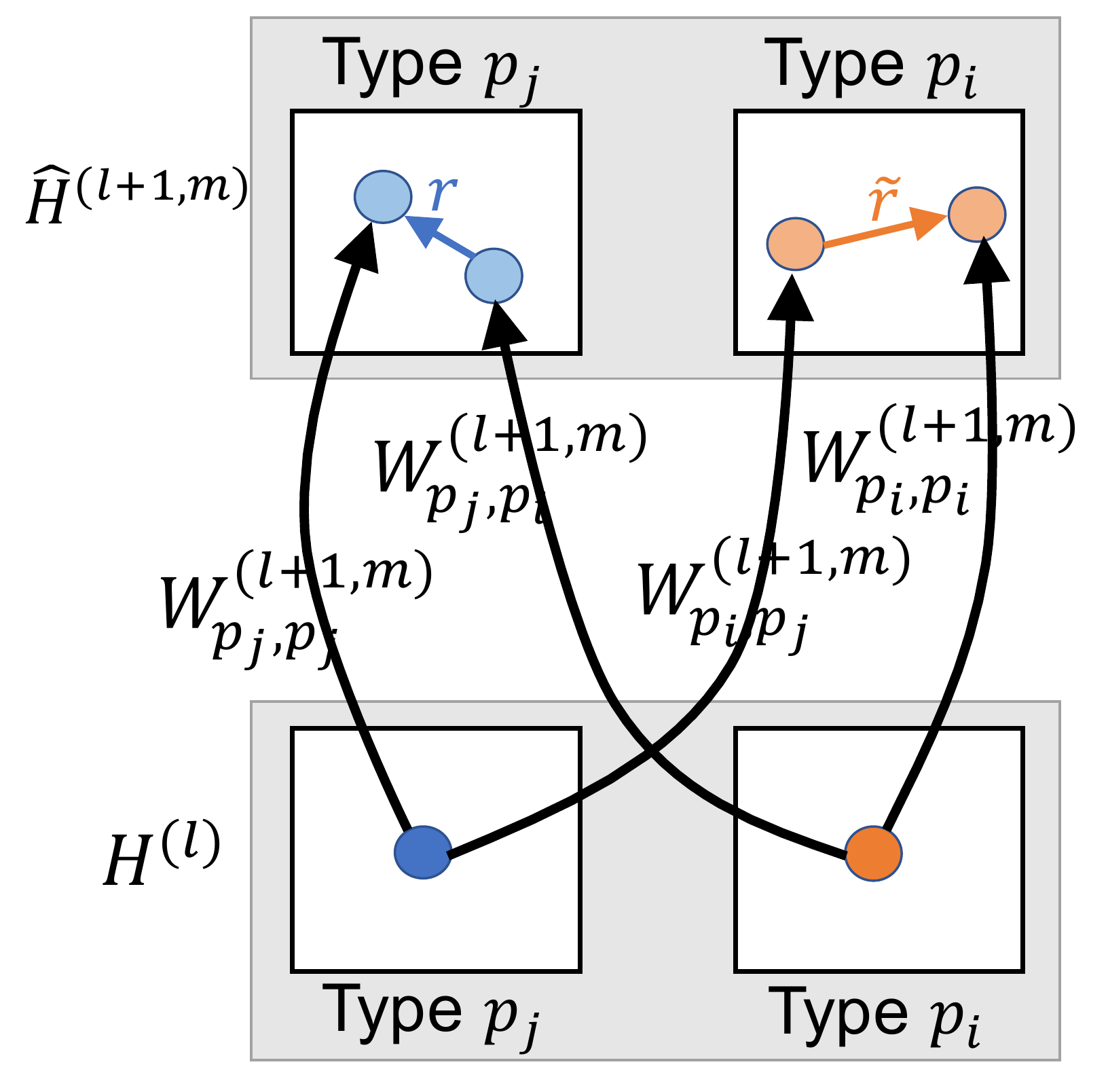}}
	\subcaptionbox{\label{fig:cycle_consis}}
	[.48\linewidth]{\includegraphics[width=0.22\textwidth]{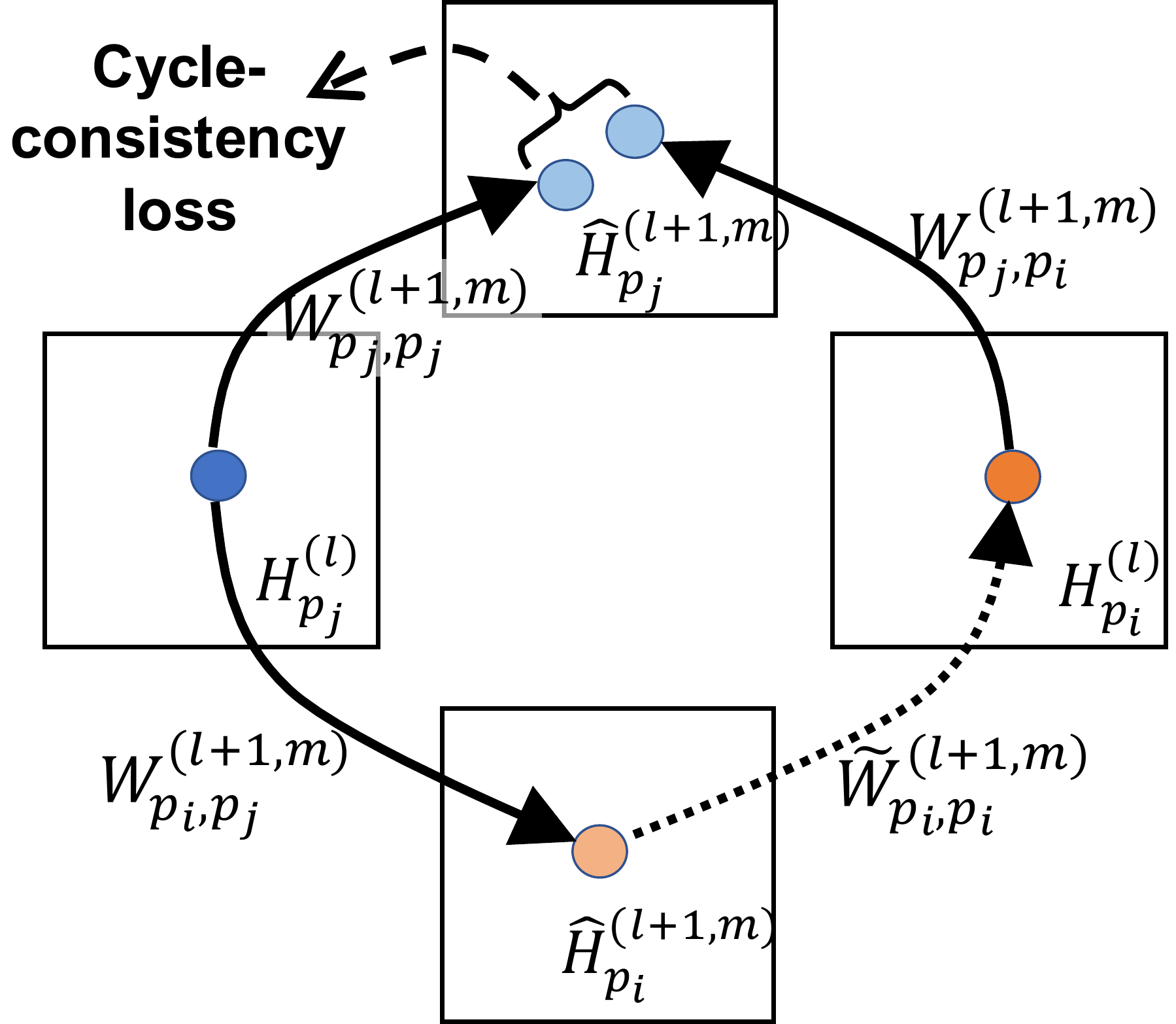}}
	\caption{An illustration for the transformation operation. (a) There are two forms of transformation between node type $p_i$ and $p_j$, i.e. $p_i \!\rightarrow\! p_j$ and $p_j\! \rightarrow \!p_i$. (b) Cycle-consistency loss.}
	\label{fig:trans_illus}
\end{figure}
\subsubsection{E1: \textit{Multi-task Learning}}
We can further employ several node classifiers for different types of nodes. The parameters of all type-aware attention layers are shared and trained by multiple classifiers. The multi-task learning via uniting all classifiers greatly reduces the risk of overfitting and benefits the robustness of representations \cite{multi_task}.

\subsubsection{E2: \textit{Voices-sharing Product}}
The \textit{concat product} scoring function considers the directed edge (e.g., write) and the reversed edge (e.g., written) as independent types of relationships. Intuitively, vertex $i$ will link to vertex $j$ with the ``written'' relation when vertex $j$ link to vertex $i$ with the ``write'' relation. To formulate the pairwise relationship between the directed edge and the reversed edge,
we share the parameters of attention mechanism between the pairwise edge types $r$ and $\tilde{r}$, where $\tilde{r}$ is the type of reversed edge of edges with type $r$.
Technically, we enforce  $\mathbf{a}_{\tilde{r}}^{\left(l,m\right)} = - \mathbf{a}_r^{\left(l,m\right)} \in \mathbb{R}^{n^{(l,m)}_{\phi(j)}}$ and adapt the attention scoring function $f_r^{(l,m)}$ as follow (called \textit{voices-sharing product}):
\begin{equation}
f_r^{(l,m)}(e) = {\mbox{$\hat{\mathbf{h}}$}_{\phi\left(j\right), j}^{\left(l,m\right)}}^{\mathsf{T}} \left(\hat{\mathbf{h}}_{\phi\left(j\right), i}^{\left(l,m\right)} + {\mathbf{a}_r^{\left(l,m\right)}}\right).
\end{equation}

\subsubsection{E3: \textit{Cycle-consistency Loss}}
In natural language processing, ``back translation and reconciliation'' has been a popular trick to verify and improve the performance on translation \cite{Back_translation}. Referring back to the transformation operation between heterogeneous nodes in Eq. \eqref{fun:trans}, we have a transformation from node type $p_i \in \mathcal{A}$ to $p_j \in \mathcal{A}$ and another transformation from $p_j$ to $p_i$. Particularly, a self-transformation is applied to per type of node, i.e. $p_i \rightarrow p_i$. The transformation operation between $p_i$ and $p_j$ is illustrated in Fig. \ref{fig:trans_oper}, which is intuitive that a vertex should return to the starting position after a cycle. Therefore, we introduce a cycle-consistency restriction to the transformation operation:
\begin{equation}
\resizebox{.88\hsize}{!}{
$W_{p_j,p_i}^{\left(l+1,m\right)} \left(W_{p_i,p_i}^{\left(l+1,m\right)}\right)^{-1} W_{p_i,p_j}^{\left(l+1,m\right)} h_j^{(l)} \approx W_{p_j,p_j}^{\left(l+1,m\right)} h_j^{(l)}$}
\label{fun:cyc_consis}
\end{equation}
where ${W_{p_i,p_i}^{\left(l+1,m\right)}}^{-1}$ is the inverse of $W_{p_i,p_i}^{\left(l+1,m\right)}$. However, the solution of matrix inversion is a notoriously time-consuming problem. To reduce the computational complexity, we adopt a trainable matrix $\widetilde{W}_{p_i,p_i}^{\left(l+1,m\right)}$ instead of the inverse of the matrix $W_{p_i,p_i}^{\left(l+1,m\right)}$ and restrain it as follows:
\begin{equation}
\widetilde{W}_{p_i,p_i}^{\left(l+1,m\right)}{W}_{p_i,p_i}^{\left(l+1,m\right)} \approx {W}_{p_i,p_i}^{\left(l+1,m\right)}\widetilde{W}_{p_i,p_i}^{\left(l+1,m\right)} \approx I,
\label{fun:cyc_inverse}
\end{equation}
where $I$ is the identity matrix. These constraints are integrated as \textit{cycle-consistency loss} (as shown in Fig. \ref{fig:cycle_consis}):
{\small{
\begin{equation}
\begin{aligned}
&\mathcal{J}_{\text{cycle}} =\\
& \beta_1\underset{p_i,p_j \in \mathcal{A}}{\mathbb{E}}\left[\left(W_{p_j,p_i}^{\left(l,m\right)} \widetilde{W}_{p_i,p_i}^{\left(l,m\right)} W_{p_i,p_j}^{\left(l,m\right)} h_j^{(l-1)} - W_{p_j,p_j}^{\left(l,m\right)} h_j^{(l-1)}\right)^2\right]\\
&+ \beta_2\underset{p\in \mathcal{A}}{\mathbb{E}}\left[\left(\widetilde{W}_{p,p}^{\left(l,m\right)}W_{p,p}^{\left(l,m\right)}-I\right)^2 + \left(W_{p,p}^{\left(l,m\right)}\widetilde{W}_{p,p}^{\left(l,m\right)}-I\right)^2\right]
\end{aligned}
\label{fun:cycle_loss}
\end{equation}}}
where $\beta_1$ and $\beta_2$ are the weighting factors. The objective function of our model therefore is derived as:
\begin{equation}
\min \mathcal{J} = \mathcal{J}_{\text{class}} + \mathcal{J}_{\text{cycle}}.
\end{equation}

\section{Experiments}\label{sec:expe}
\paragraph{Comparative Models} The list of models in comparison includes:

\textbf{{1) Variants of our proposed model\footnote{Available at \url{https://github.com/didi/hetsann}}:}} We denote HetSANN as the proposed vanilla version, i.e. without aforementioned three extensions in Section \ref{sec:3extension}. Three suffixes ``.$M$'', ``.$R$'' and ``.$V$'' indicate \textit{multi-task learning} to optimize the parameters, \textit{voices-sharing product} in relations attention mechanism and \textit{cycle-consistency loss} to retain the transformation between vertices, respectively. And HetSANN.$M$.$R$.$V$ refers to the full version of our proposed model.

All variations employ 3-layer HetSANN and each TAL consists of 8 attention heads. The output dimensions of each attention head are consistent to 8. The parameters are optimized via Adam solver \cite{adam} with a learning rate 0.001 for IMDB and 0.005 for other datasets. A regularization weight 0.0005 is applied to all trainable parameters. 
A dropout rate 0.6 \cite{dropout} is implanted between hidden layers to stabilize our model training procedure. For the variant of HetSANN with suffix ``.$V$'', the weight coefficients $\beta_1\! = \!10^{-3}$ and $\beta_2 \!=\!10^{-5}$.

\textbf{{2) Baseline models:} } We compare with the state-of-art baselines of which codes is publicly available at the website, including DeepWalk \cite{deepwalk}, metapath2vec \cite{metapath2vec}, HERec \cite{HERec}, HAN \cite{HAN}, GCN \cite{gcn}, R-GCN \cite{rgcn} and GAT \cite{GAT}. All of baseline models are introduced in Section \ref{sec:related}, and the implementation of them is detailed as follows.

We turn the homogeneous graph embedding methods (DeepWalk, GCN and GAT) into the model for the heterogeneous graph embedding learning by ignoring the type of nodes and linkages. R-GCN is implemented by regarding all nodes as the same type node but distinguishing different types of relations in the graph. We follow most of parameters setting recommended in the published papers, and tune a few parameters to adapt to the dataset in our experiments. Specifically, we set the walk length to 50 and 100 walks/node for metapath2vec, HERec and DeepWalk, and the learned embeddings by them are used to train a 2-layer MLP \cite{MLP} classifier. For the graph neural network based methods, the number of layers is set to 3 for GCN, R-GCN and GAT. We use 8 attention heads in each layer of the attention-based models, i.e., HAN and GAT. To enable comparison, we design some meta-path schemes for each dataset and evaluate all meta-paths schemes for metapath2vec and HAN, and report the best performance.

All networks were trained from scratch until convergence. The dimension of the embedding is unanimously set to 64 for all models. And the model settings for all dataset are the same except for special instructions.
\begin{table}[t]
\centering
\caption{The statistics of datasets in our experiments.}\smallskip
\label{tab:datasets}
\resizebox{0.48\textwidth}{!}{
\begin{tabular}{clll}
\toprule
Dataset& Nodes & Linkages & isLabel\\
\midrule
\multirow{3}{*}{IMDB}& movie(5043) & movie-actor(11188)& movie\\
& actor(2357)& movie-director(3435)&\\
& director(894)&&\\
\midrule
\multirow{3}{*}{DBLP}& author(14475) & author-paper(41794)& author\\
& paper(14376)&paper-venue(14376)& paper\\
& venue(20) & &\\
\midrule
\multirow{3}{*}{AMiner}& author(8052)&author-author(31224)& author\\
& paper(20201)& paper-paper(44551)& paper\\
& &author-paper(32029)&\\ 
\bottomrule
\end{tabular}}
\end{table}
\begin{table*}[t]
\centering{}
\caption{Comparison results for node classification in datasets.}\smallskip
\label{tab:compar_results}
\begin{tabular}{l|rr|rr|rr|rr|rr}
\toprule
Dataset&	\multicolumn{2}{c|}{IMDB}&		\multicolumn{4}{c|}{DBLP}&				\multicolumn{4}{c}{AMiner}\\		
Target node&	\multicolumn{2}{c|}{Movie}&		\multicolumn{2}{c}{Author}&		\multicolumn{2}{c|}{Paper}&		\multicolumn{2}{c}{Author}&		\multicolumn{2}{c}{Paper}\\
Metrics (\%)&	Mic F1&	Mac F1& Mic F1&	\multicolumn{1}{r}{Mac F1}&	Mic F1&	Mac F1&	Mic F1&	\multicolumn{1}{r}{Mac F1}&	Mic F1&	Mac F1\\
\midrule
DeepWalk&	63.53&	54.91&	92.71&	92.01&	99.41&	99.30&	84.70&	84.99&	87.71&	87.70\\
metapath2vec&	60.83&	50.27&	66.75&	67.07&	70.35&	72.89&	61.79&	61.72&	71.93&	71.58\\
HERec&	62.54&	53.62&	90.49&	89.94&	99.93&	99.93&	68.74&	68.77&	78.75&	78.61\\
HAN&	61.91&	57.87&	88.35&	87.67&	\textbf{100.00}&	\textbf{100.00}&	33.24&	31.92&	91.47&	91.73\\
GCN&	63.78&	51.13&	87.09&	86.60&	91.31&	89.87&	81.24&	81.69&	90.55&	90.76\\
R-GCN&	67.13&	62.58&	87.70&	86.96&	81.92&	79.37&	85.11&	85.31&	90.84&	91.05\\
GAT&	65.19&	60.43&	89.11&	88.57&	89.14&	87.51&	86.86&	87.44&	90.79&	90.96\\
\midrule
\midrule
HetSANN&	\multirow{2}{*}{73.11}&	\multirow{2}{*}{71.20}&	94.89&	94.56&	99.72&	99.67&	86.97&	87.48&	91.20&	91.37\\
HetSANN.$M$&	&	&	95.43&	95.21&	99.08&	98.89&	91.55&	91.99&	92.56&	92.77\\
HetSANN.$M$.$R$&	73.20&	71.38&	95.51&	95.28&	99.11&	98.95&	92.43&	92.87&	\textbf{93.75}&	\textbf{93.93}\\
HetSANN.$M$.$R$.$V$&	\textbf{73.86}&	\textbf{72.00}&	\textbf{95.63}&	\textbf{95.38}&	98.69&	98.43&	\textbf{92.47}&	\textbf{92.91}&	93.73&	93.92\\
\bottomrule
\end{tabular}
\end{table*}
\paragraph{Datasets}We collected a movie graph from IMDB site\footnote{\url{https://www.imdb.com}} and constructed two academic networks from DBLP \cite{dblp_dataset} and AMiner \cite{arnetminer} datasets respectively. Each of these three data sets, the statistics of which are tabulated in Table \ref{tab:datasets}, is a heterogeneous graph, consisting of more than two types of nodes or edges.
\begin{itemize}
	\item \textbf{IMDB} The IMDB records the Actors and Directors of the Movies. The movies are divided into three groups according the genre label: Action, Comedy or Drama.
	We utilize the keywords about plot of the movie as the attribute feature of movie vertex by the way of bag-of-words.
	For meta-path-based models, we set two meta-path schemes, i.e., \textit{MAM} and \textit{MDM}.

	\item \textbf{DBLP} In an academic network, Authors published their Papers in the Venues. The DBLP dataset we constructed consists of 20 venues from four different research fields: database, data mining, machine learning, and information retrieval. Each paper is labeled according to the research field of the venue where the paper is published, and each paper is characterized by the bag-of-words of keywords.
	Each author is labeled based on the research fields of her/his publications, and we sum up the bag-of-words of the papers published by this author as the feature of the author. Again, we set \textit{PAP} and \textit{PVP}, \textit{APA} and \textit{APVPA} as the meta-path schemes for meta-path-based models in paper classification task and author classification task respectively.

	\item \textbf{AMiner} We cast aside venue nodes from AMiner academic network, raising a harder classification task.
	In addition to the publishing relationship between the paper and the author, we also introduce the citation relationship between papers and the  collaboration relationship between authors. Similar to DBLP, each paper in AMiner is characterized by the bag-of-words of keywords, and papers and authors are labeled into four research fields: database, data mining, natural language processing and computer vision.
	The attribute features of each author is provided with five indices\footnote{Five indices \cite{5indices} for each author: the count of published papers; the total number of citations; the H-index; the P-index with equal A-index; the P-index with unequal A-index.} indicating the academic authority of the author. Again, \textit{PAP} and \textit{PAAP}, and \textit{APA} and \textit{APPA} are set as meta-path schemes.
\end{itemize}
\paragraph{Evaluation Metrics}
The whole labeled dataset is randomly split into training set, validation set and test set by a ratio of 0.8:0.1:0.1.
And we select the best one in the validation set for each comparative model, then evaluate them by Micro F1 and Macro F1 on the test set. For each model, we report the average performance on 10 repeated processes.

\subsection{Ablation Study}
In this section, we employ the vanilla HetSANN and its variations including HetSANN.$M$, HetSANN.$M$.$R$ and our full method HetSANN.$M$.$R$.$V$ to perform an ablation study. To enable the multi-task learning, we introduce the Paper classification as the auxiliary task to the main Author classification task. And the \textit{multi-task learning} (with suffix ``.$M$'') can not be conducted on the IMDB dataset which contains only single type of labeled node. The test results are shown in the bottom half of Table \ref{tab:compar_results}: (1) HetSANN.$M$ performs better than HetSANN both in Author and Paper classification tasks in AMiner. For DBLP dataset, HetSANN.$M$ still brings improvement in the task of Author classification, although it is slightly inferior in the performance of the Paper classification compared to HetSANN that is trained toward only one single task. We believe that the introduced \textit{multi-task learning} guides our model to find an optimum representation that captures all of tasks, even if sometime it involves losing accuracy of one task in return for gaining performance in overall \cite{multi_task};
(2) Benefiting from the substitution of \textit{concat product} with \textit{voices-sharing product}, HetSANN.$M$.$R$ improves the classification performance over all datasets compared with HetSANN.$M$;
(3) HetSANN.$M$.$R$.$V$ achieves the best performance of variants of our models in the Author and Movie classification tasks. However, the gain of HetSANN.$M$.$R$.$V$ is relatively constrained as seen from the comparison between HetSANN.$M$.$R$ and HetSANN.$M$.$R$.$V$. One clear reason is the replacement of the analytical expression of the inverse matrix by the trainable matrix in the cycle-consistency loss [Eq. \eqref{fun:cycle_loss}], which is left to future work.

\subsection{Comparison Results}
Table \ref{tab:compar_results} also shows the comparison results of our models with other baselines. Obviously, our models are superior to other models in the classification tasks of all datasets excluding Paper classification on DBLP. Note that we label papers according to the research field of the venue where the paper is published. The venue nodes connected to papers in DBLP enable HAN to establish the neighborhood of papers published in the same venue via the meta-path scheme \textit{PVP}, resulting in a perfect Paper classification performance on DBLP. Without the venue vertices in AMiner, it is not easy for HAN to capture the category information of Paper via \textit{PAP} and \textit{PAAP} schemes, leading to worst Author classification results in AMiner.
Different from the methods which require well-designed solutions of the meta-path, the meta-path-free methods achieve obvious performance gains and robustness results. 
With a further distinction between multiple types of nodes and linkages, our models outperform other baselines and
our full method HetSANN.$M$.$R$.$V$ improves Micro F1 and Macro F1 by $3\%\!\sim\!13\%$ and $3\%\! \sim\! 19\%$ respectively over the most competitive model GAT on three datasets.

\begin{figure}[ht]
	\centering
	\includegraphics[width=.98\columnwidth]{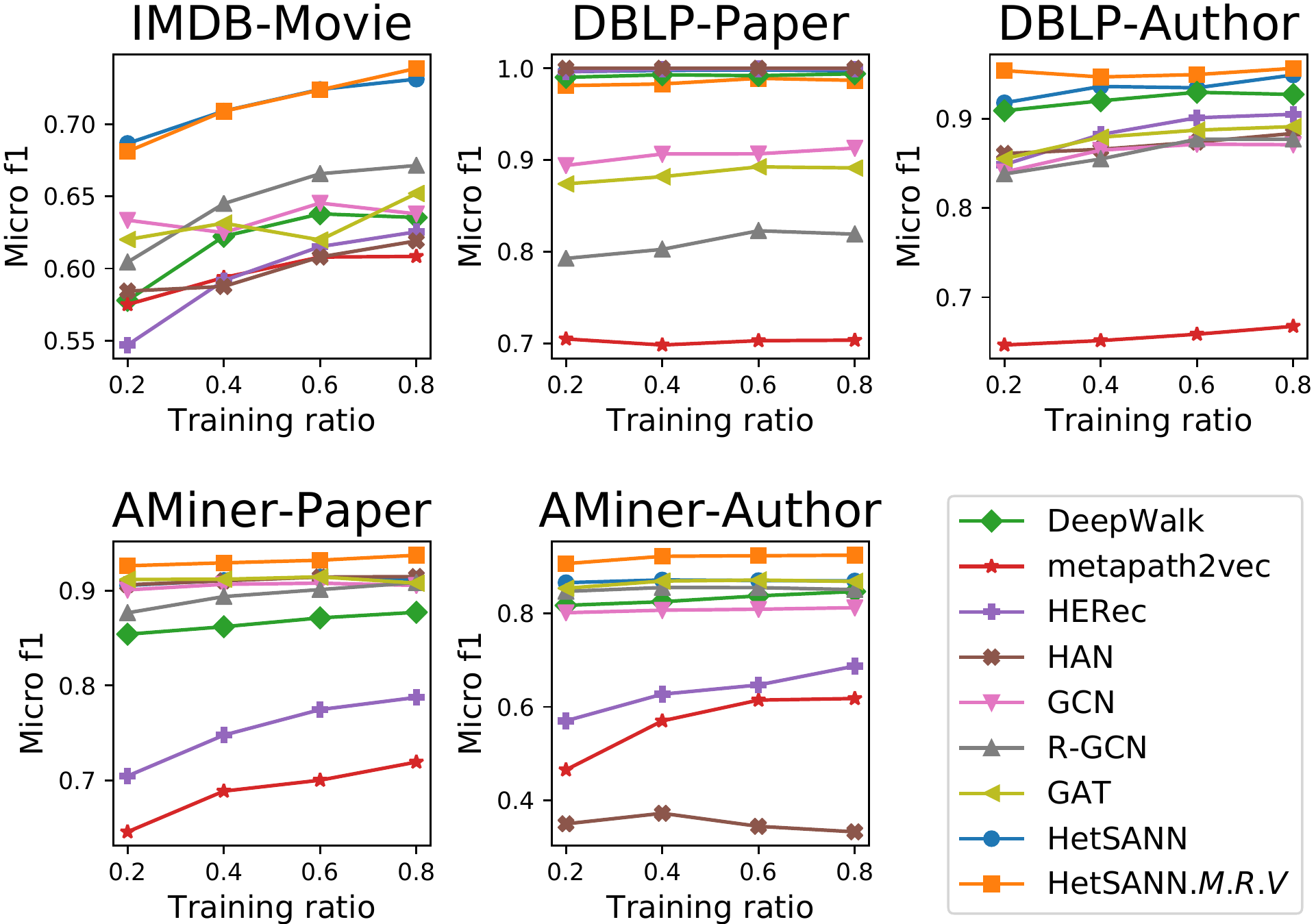}
	\caption{Micro F1 vs. Training ratio.}
	\label{fig:training_ratio}
\end{figure}

Fig. \ref{fig:training_ratio} details the comparison performance results of main tasks (Movie and Author classification) and the auxiliary tasks (Paper classification) on the three datasets, varying the value of training ratio in $\{0.2,\! 0.4,\! 0.6,\! 0.8\}$. Consistently, we have HetSANN.$M$.$R$.$V$$>$HetSANN$>$baselines in terms of Micro F1 score of classification. Besides, both HetSANN and HetSANN.$M$.$R$.$V$ still maintain the advantage in a weakly-supervised manner.

\subsection{Parameter Sensitivity Study}
\begin{figure}[t]
	\centering
	\subcaptionbox{Sensitivity to $L$}
	{\includegraphics[width=0.32\columnwidth]{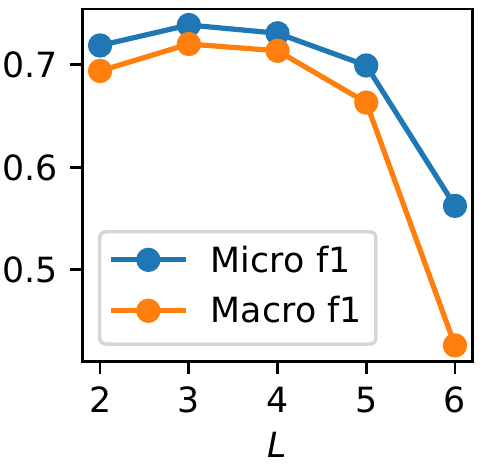}}
	\subcaptionbox{Sensitivity to $\beta_1$\label{fig:beta1}}
	{\includegraphics[width=0.33\columnwidth]{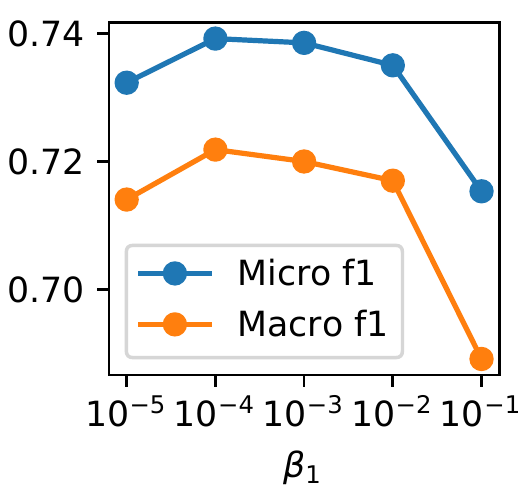}}
	\subcaptionbox{Sensitivity to $\beta_2$\label{fig:beta2}}
	{\includegraphics[width=0.33\columnwidth]{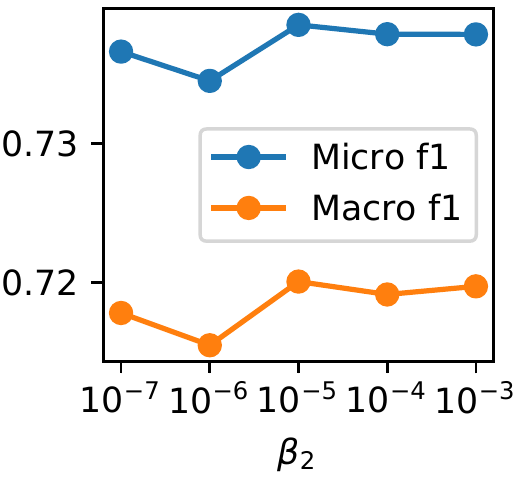}}
	\caption{Parameter analysis of HetSANN.$R$.$V$ on IMDB.}
	\label{fig:parameters}
\end{figure}

Finally, we test the parameter sensitivity of HetSANN.$R$.$V$ on IMDB and the results are presented in Fig. \ref{fig:parameters}. The left figure shows the effect of the number of type-aware attention layers $L$ when other parameters are fixed. The performance of HetSANN.$R$.$V$ goes down when $L\!>\!5$. The observation is consistent with the analysis in \cite{GCNdeeper}, that is, the graph convolution is a form of Laplacian smoothing over the features of neighborhood, and it will lead to indistinguishable features of the node and inferior performance of node classification when many convolutional layers are included in a GNN. The remaining two figures focus on the weighting factors $\beta_1$ and $\beta_2$ of the cycle-consistency loss. Fixing $\beta_2\!=\!10^{-5}$, the lines in Fig. \ref{fig:beta1} increase when $\beta_1$ increases to $10^{-4}$ and lines decrease when we set a larger $\beta_1$, which may suppress the learning of main task, i.e., node classification. Fig. \ref{fig:beta2} in turn varying $\beta_2$ and fixing $\beta_1\!=\!10^{-3}$, lines tend to be stable when $\beta_2\!>\!10^{-5}$, indicating that the model have done its best to reach the solution of the inverse matrix in Eq. \eqref{fun:cyc_inverse}.

\section{Related Work}\label{sec:related}
\subsection{Heterogeneous Graph Representation Learning}
The existing works of HIN embedding tend to utilize the meta-path to adapt the heterogeneous graph for the application of the homogeneous graph embedding methods, such as \cite{deepwalk,LINE,SDNE}.
metapath2vec \cite{metapath2vec} designs meta-paths to guide the random walks in a heterogeneous graph and then follows skip-gram model to learn the latent-space representations of vertices. Inspired by metapath2vec, HERec\cite{HERec} proposed to fuse different representations learned in the view of different meta-path schemes. Both metapath2vec and HERec are trained by linkage-guided objective function which is independent of the downstream tasks. To obtain optimum embeddings for a specific task, \cite{chen2017task} joints author identification task-guided and linkage-guided objectives to learn the heterogeneous graph embeddings.
HAN \cite{HAN} introduces a two-levels hierarchical attention to GNN, in which a node-level attention captures the relations between neighboring nodes generated by one meta-path scheme and a semantic-level attention aggregates multiple meta-path scheme for each node in a graph. All in all, these aforementioned methods are dependent on the design of experts and brings inevitable loss of information.

\subsection{Graph Neural Networks (GNNs)}
More recently, graph neural networks (GNNs) \cite{GNN,graphSAGE,gcn,GN} have become increasingly studied. GNNs generate node embeddings by the spatial filter which convolutes each node over its neighborhood in graph. The convolutional operation enables GNNs to propagate structural information of graphs layer by layer, and frees graph embedding methods from linkage-guided learning. Motivated by the thriving of attention mechanism, GAT \cite{GAT} introduces an attention strategy to GNN framework. For the relational learning of knowledge bases, R-GCN \cite{rgcn} builds multiple relation spaces for all nodes in a graph, which can not capture the relative importance of various type of nodes. RSHN \cite{RSHN} focuses on mining semantic interactions of edge types via coarsened line graph and incorporating it into the H-GNN model.
\section{Conclusion}\label{sec:conclu}
The paper proposes HetSANN to perform meta-path-free embedding based on structural information in heterogeneous graphs. We design a type-aware attention layer for HetSANN, which embeds each vertex of heterogeneous graph by jointing different types of neighboring nodes and associated linkages. A few variants of our model are developed based on three extensions, i.e., \textit{voices-sharing product}, \textit{cycle-consistency loss} and \textit{multi-task learning}. Comprehensive experiments on three popular datasets show that the proposed solutions outperform state-of-the-art methods in HIN embedding and node classification. Under the framework of HetSANN, the representation learning of HIN does not need to rely on the meta-path to tackle the heterogeneous structural information, thereafter the heterogeneous attributes of vertices will be considered in the future work.

\bibliography{AAAI-HongH.5327}
\bibliographystyle{aaai}
\end{document}